\pdfoutput=1
\PassOptionsToPackage{prologue,dvipsnames}{xcolor}

\documentclass[11pt]{article}

\usepackage[]{acl}

\usepackage{times}
\usepackage{latexsym}
\usepackage{amssymb}
\usepackage{graphicx}
\usepackage{amsmath}
\usepackage{algorithm}
\usepackage{algpseudocode}
\usepackage{tabularx}
\usepackage{array}
\newcolumntype{I}{!{\vrule width 3pt}}
\usepackage{multirow}
\usepackage{diagbox}
\usepackage{listings}
\usepackage{caption}

\usepackage[T1]{fontenc}

\usepackage[utf8]{inputenc}

\usepackage{microtype}

\usepackage{inconsolata}
\usepackage{enumitem}

\usepackage{blindtext}
\newcommand\blfootnote[1]{%
\begingroup
\renewcommand\thefootnote{}\footnote{#1}%
\addtocounter{footnote}{-1}%
\endgroup
}

\definecolor{blue1}{RGB}{227, 239, 255} 
\definecolor{blue2}{RGB}{198, 219, 255}
\definecolor{blue3}{RGB}{169, 199, 255}
\definecolor{blue4}{RGB}{140, 179, 255}
\definecolor{blue5}{RGB}{111, 159, 255}
\definecolor{blue6}{RGB}{82, 139, 255}
\definecolor{blue7}{RGB}{53, 119, 255}
\definecolor{blue8}{RGB}{24, 99, 255}
\definecolor{blue9}{RGB}{0, 85, 255}
\definecolor{blue10}{RGB}{0, 56, 255} 

\definecolor{orange1}{RGB}{255, 235, 215}
\definecolor{orange2}{RGB}{255, 225, 195}
\definecolor{orange3}{RGB}{255, 215, 175}
\definecolor{orange4}{RGB}{255, 205, 155}
\definecolor{orange5}{RGB}{255, 195, 135}
\definecolor{orange6}{RGB}{255, 185, 115}
\definecolor{orange7}{RGB}{255, 175, 95}
\definecolor{orange8}{RGB}{255, 165, 75}
\definecolor{orange9}{RGB}{255, 155, 55}
\definecolor{orange10}{RGB}{255, 145, 35}

\title{Do Clinicians Know How to Prompt? The Need for Automatic Prompt Optimization Help in Clinical Note Generation}

\author{Zonghai Yao \thanks{* Indicates equal contribution} $^1$, Ahmed Jaafar\footnotemark[1] $^1$, Beining Wang $^2$, Zhichao Yang $^1$, \bf{Hong Yu}$^{1}$\\
University of Massachusetts Amherst$^1$, Fudan University$^2$\\
{\tt \{zonghaiyao, ajaafar\}@umass.edu}\\ 
}

\begin{document}

\maketitle

\begin{abstract}

This study examines the effect of prompt engineering on the performance of Large Language Models (LLMs) in clinical note generation. We introduce an Automatic Prompt Optimization (APO) framework to refine initial prompts and compare the outputs of medical experts, non-medical experts, and APO-enhanced GPT3.5 and GPT4. Results highlight GPT4-APO's superior performance in standardizing prompt quality across clinical note sections. A human-in-the-loop approach shows that experts maintain content quality post-APO, with a preference for their own modifications, suggesting the value of expert customization. We recommend a two-phase optimization process, leveraging APO-GPT4 for consistency and expert input for personalization
~\footnote{\url{https://github.com/seasonyao/Automatic_Prompt_Optimization_Physician_Prompting}}.

\end{abstract}

\section{Introduction}
\blfootnote{To appear in BioNLP 2024}

\begin{figure}[t]
\centering
\includegraphics[width=0.45\textwidth]{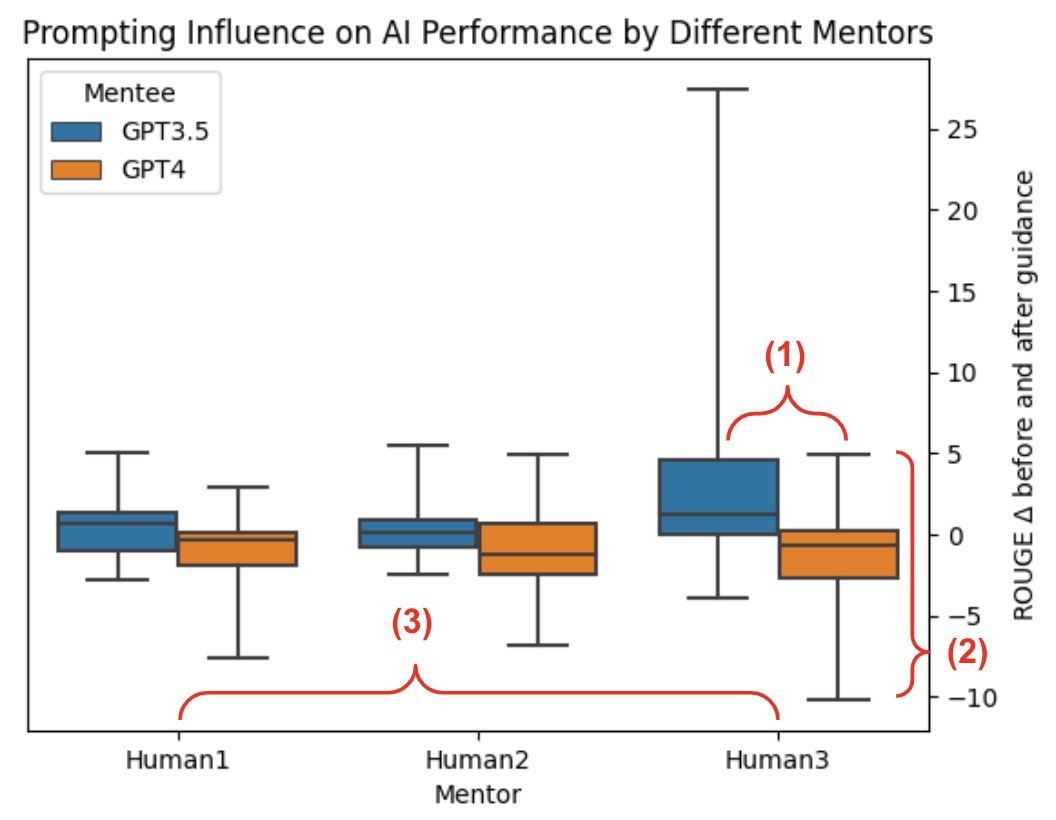}
\vspace{-4mm}
\caption{Influence of different mentors on AI mentee performance enhancement. This figure illustrates the changes in AI mentee performance following prompting by three individual human mentors and an APO system, represented on the x-axis. The y-axis measures the variation in ROUGE scores before and after prompting, with blue bars indicating GPT3.5 and orange bars denoting GPT4 as mentee to generate clinical note content according to different prompt groups. The results indicate the differential impact of human versus APO prompting on AI content generation quality.}
\label{fig:mentor_guide_mentee}
\vspace{-4mm}
\end{figure}

Large Language Models (LLMs), including iterations of the Generative Pre-trained Transformer (GPT) series, have dramatically expanded the scope of natural language processing (NLP). 
Their applications now range from simple Q\&A to the intricate demands of clinical documentation, necessitating the craft of prompt engineering~\cite{brown2020language,Sanh2021,chowdhery2022palm,longpre2023flan,openai2023gpt4,wang2023prompt,Yang2023.10.26.23297629}. 
The quality of a prompt is paramount, as it is typically created by a human mentor to guide an LLM mentee to generate the document. 
Yet, this prompt creation process is encumbered by the complexities of human expression—rich in subtleties and cultural nuance—that often surpass the computational confines of LLMs, resulting in a cognitive gap~\cite{zamfirescu2023johnny}.
Variances in prompt quality lead to differences in prompt efficacy, which can fluctuate considerably 
\textcolor{red}{(1)} when switching between LLM mentees (As shown in Figure~\ref{fig:mentor_guide_mentee}, `mentor' modifies the prompt to allow `mentee' to perform the targeted task better) and 
\textcolor{red}{(2)} across various sections of the documentation or 
\textcolor{red}{(3)} among different human mentors, as illustrated in Figure~\ref{fig:mentor_guide_mentee}.
This inherent variability underscores the need for a consistent tool that standardizes prompt quality to achieve reliable uniformity in LLM performance.

In the clinical domain, where the stakes are particularly high, optimizing prompt engineering is critical to help busy clinicians most efficiently use LLMs for clinical practice.
Our study adopts Automatic Prompt Optimization (APO)~\cite{prasad2022grips} as a novel solution to address these challenges.
APO refines the initial prompts provided by clinicians, adapting them to the nuanced requirements of different clinical note sections for AI-assisted clinical documentation. Thus, the resulting clinical notes are significantly enhanced in quality and efficiency.

Through a comprehensive comparative analysis, our research elucidates how APO, when used with human experts, substantially elevates the refinement process of prompts.
Our first experimental set pits generic prompts, modified by medical experts, non-medical experts, and APO-enhanced GPT3.5 and GPT4, against each other. 
The results highlight APO-GPT4's remarkable ability to elevate content generation, revealing an inherent capacity for self-improvement that aligns with recent academic discourse.
Our second experimental set delves into the potential of human-in-the-loop systems. 
Here, we further refine APO-generated prompts with human experts. 
Contrary to non-expert interventions, which often detracted from the quality of the content, expert modifications maintained the high standards set by APO. 
Moreover, our human preference feedback suggests that, while experts may not significantly alter the content quality, they prefer the results of their own modifications, pointing to a personalized touch without sacrificing the quality of the content.

In light of our findings, we advocate a two-pronged approach to prompt optimization: 
initially employing APO-GPT4 to standardize prompt quality, 
followed by expert-led customization based on preference. 
This strategy offers a pragmatic balance, effectively harnessing the power of AI while respecting the nuances of human expertise.

\section{Related Work}
Soft prompts and parameter adjustments offer promising results for open-source LLMs \cite{li2021prefix, lester2021power, hu2021lora}, while discrete prompt searches \cite{shin2020autoprompt, wen2023hard} and reinforcement learning \cite{deng2022rlprompt, zhang2022tempera} push the boundaries further. Closed-source LLMs, conversely, necessitate gradient-free optimization, relying on iterative prompt refinement and natural language feedback for efficacy~\cite{prasad2022grips, xu2022gps, guo2023connecting, fernando2023promptbreeder, zhou2022large, xu2023reprompting, pryzant2023automatic, yang2023large, wang2023promptagent, dong2023pace, li2023automatic, sun2023autohint}.

In the clinical context, synthesizing such optimization techniques has been pivotal. Foundational work in automated note generation \cite{krishna2020generating, song-summarizing, yim2021towards, Su2022ExtractAA, giorgi2023wanglab, wang2023umass_bionlp, wang2023notechat, yao2023improving} informs our approach, integrating APO to streamline medical documentation. This research leverages both iterative enhancement and expert feedback, embodying the iterative, gradient-free optimization approach to improve the precision of clinical LLM applications.

\section{Method}

We are given a dataset $D$ of $n$ i.i.d training clinical data, comprised of $f$ features ($D \in \mathbb{R}^{n \times f}$)
including the doctor-patient dialogue, the name of a SOAP ~\cite{podder2021soap, soap} note section~\footnote{SOAP structure details can be found in the Appendix \ref{SOAP-Structure-appendix}.}, the ground truth section clinical note summary, the model-generated section clinical note summary, etc.
Our method broadly consists of a ``forward pass'' (\ref{sec:forward}) and a ``backward pass'' (\ref{sec:backward}).
First, an LLM generates summaries for a batch $h$ from a section $s \in S$ using a generic prompt $p_0$ provided by the user.
An LLM is then asked via a fixed prompt $p_\nabla$ to provide suggestions to make $p_0$ more suitable for $s$ given the ground truth and generated summaries, producing an answer $g$.
Afterward, another fixed prompt, $p_\delta$, is used to command the LLM to use $g$ to fix $p_0$, outputting a new prompt $p'$.
$p'$ should now be slightly more tailored to generate better summaries for $s$, closer to the theoretical optimal prompt $p^*$.
This is executed for all $S$ utilizing a random sample of data $h$ (batch) from each section, where $h \subseteq n$.
This process is illustrated in Figure \ref{fig:flow} and detailed in Algorithm \ref{algo}~\footnote{Algorithm \ref{algo} is simplified to use one data point's dialogue ($x$). In reality, a batch ($h$) of data is used.
Note that iterations for batch h involve a single type but not multiple types of sections.}.

\subsection{Forward Pass}\label{sec:forward}
The forward pass utilizes an LLM to generate summaries ($\hat{y}$) for $h$ from section $s$ by passing in a generic user-provided prompt ($p_0$), doctor-patient dialogue ($x$), and $s$.
We use black box LLMs via API, denoted as $LLMp(i)$~\footnote{$i$ is defined as all the inputs to the prompt (dialogue, section, etc.).}.
This API yields a probable text continuation, symbolized as $\hat{y}$, given a prompt. 
This prompt is a fusion of $p$ and $i$. 
Mathematically, $LLMp(i)$ is approximated by $\text{argmax}_{\hat{y} \in L} P_{\text{LLM}} (\hat{y} | p, i)$, where it selects the most likely continuation $\hat{y}$ from the set of natural language tokens $L$.
The ones used for our method are OpenAI's \textbf{GPT3.5} and \textbf{GPT4}~\footnote{We use OpenAI's \texttt{gpt-3.5-turbo-0613} and \texttt{gpt-4-0613} in our experiments.}.

$p_0$ is a generic prompt such as the one shown in Figure \ref{fig:flow} or Appendix \ref{prompts} that, in our use case, would be provided by a medical professional such as a clinician.
It is a prompt that only instructs the model, in this step LLM $a$.
$p_0$ and $x$ are passed into $a$ to output a generated summary $\hat{y}$.
This first $\hat{y}$ is likely to be very suboptimal for $s$.

\vspace{-2mm}
\begin{algorithm}
\caption{SOAP Note Prompt Optimization}
\label{algo}
\begin{algorithmic}[1] 
\State \(p_0 = \) ``Generate a SOAP summary.''
\State \(p_\nabla = \) ``What's wrong with $p_0$?''
\State \(p_\delta = \) ``Use $g$ to fix $p_0$.''

\Procedure{forward}{$s, x$}
    \State \(p_0 = p_0 + s + x\)
    \State \Return \text{a}($p_0$) \Comment{LLM $a$}
\EndProcedure

\Procedure{backward}{$s, x, y, \hat{y}$}
    \State \(p_\nabla = p_\nabla + p_0 + s + x + y + \hat{y}\)
    \State $g =$\text{ b}($p_\nabla$) \Comment{LLM $b$}
    \State \(p_\delta = p_\delta + p_0 + g\)
    \State \Return \text{ b}($p_\delta$) \Comment{LLM $b$}
\EndProcedure

\Procedure{main}{}
     \For{\(i = 1\) \textbf{to} \(k\)}
        \For{\(c = 1\) \textbf{to} \(j\)}
            \State \(\hat{y} = \Call{forward}{x, s} \)
            \State \(p' = \Call{backward}{s, x, y, \hat{y}} \)
            \State \(p_0 = p'\)
        \EndFor
    \EndFor
\EndProcedure

\end{algorithmic}
\end{algorithm}
\vspace{-6mm}

\subsection{Backward Pass}\label{sec:backward}
This segment of the algorithm represents the key transformational stage.
The backward pass consists of (1) utilizing the same or a different LLM as before to provide suggestions on what is wrong with $\hat{y}$,  (2) utilizing the LLM in step 1 to fix $p_0$ using the suggestions provided in step 1.
Step 1 generates ``gradients'' and step 2 performs ``backpropagation''.

The backward pass starts by passing in a fixed prompt ($p_\nabla$), $p_0$, $x$, $s$, the ground truth summaries ($y$), and $\hat{y}$ into an LLM $b$ to generate suggestions ($g$) on how to fix $p_0$ to make it more suitable for generating summaries for $s$.
An example is shown in Appendix \ref{prompts}.
These suggestions are named ``gradients'', the reason $p$ is labeled with $\nabla$.
Note that $a \stackrel{?}{=} b$, i.e. $a$ may or may not be equal to $b$.

Next, a fixed prompt ($p_\delta$), like the one shown in Appendix \ref{prompts}, commands $b$ to use $g$ to fix $p_0$.
$g$, $p_0$, and $p_\delta$ are passed into $b$.
``gradient descent'' happens here.
$p_\delta$ resembles differentiation in traditional neural network training by using $g$ (the ``gradient'') to guide the model toward a lower ``loss''.
Hence the $p$ is labeled with $\delta$.
A new prompt $p'$ is outputted by $b$, which should be closer to the optimal prompt $p^*$.
$p^* = \text{argmax}_{p \in L} \{m(p, T)\}$, where $m(\cdot)$ represents a metric function and $T$ is all the training data for $s$.
$p'$ should be an edited version of $p_0$ that is in the opposite semantic direction. 

\subsection{Iterations \& Validation}
At this point in the algorithm, the same $h$ is summarized again using $a$, but this time with $p'$.
The new summaries are evaluated against $y$.

$p'$ is set to $p_0$ and the ``iteration'' restarts, repeating $j$ times.
After $j$ iterations, the ``epoch'' is finished, and the final prompt, $p'_{final}$, is used to generate summaries for a validation dataset $E$.
These summaries are evaluated against $y$ to check the performance of $p'_{final}$.
The epochs are repeated $k$ times.

\subsection{Human-in-the-Loop Prompt Refinement}\label{sec:human_loop}

Enhancing the APO framework, we incorporate a human-in-the-loop component for prompt refinement. 
Post-APO, medical experts and laypersons review and adjust $p'_{final}$ for each $s$, adding clinical acumen to the AI's output.
These revised prompts, $p'_{final-human}$, are then evaluated by generating new summaries and scoring them against ground truths.
The goal is to determine if there is a potential for human-AI collaboration on this task, and whether it should be with experts or not.


\section{Experiments}\label{sec:exp}

\subsection{Dataset} 
\label{dataset-appendix}
With 1.7k total doctor-patient dialogues and summaries, MTS-Dialog supports advances in automatic clinical note generation \cite{abacha2023empirical, abacha2023overview}.
For our initial exploration of which GPT variants are the best across most sections (more details in Section \ref{results}), we use the original evaluation split of 100 data points.
For APO, since the evaluation split is small, we merge the training and evaluation data into a single pool. 
The data is comprised of 20 SOAP sections.
We discard sections with less than 10 data points, resulting in 14 sections that meet the criteria for further experimentation.
Then, we randomly sample 5 data points from each section as training data.
Detailed data distribution for these sections is outlined in the Appendix Table~\ref{table:data_size}.

\subsection{Metrics}
Models are evaluated with full-length F1-scores of ROUGE~\cite{lin2004rouge} and METEOR~\cite{banarjee2005}. 
We use QuickUMLS\footnote{https://github.com/Georgetown-IR-Lab/QuickUMLS} to extract medical concepts from both model-generated and ground truth summaries and then calculate F1-scores for these two lists of concepts, which is named UMLS-F1~\cite{adams2023meta}.
We also add human preferences in Experiment Set-2.

\begin{table}
\centering
\scalebox{0.82}{
\begin{tabular}{c|ccccc}
\hline
\multirow{2}{*}{Mentor} & \footnotesize{R1} & \footnotesize{R2} & \footnotesize{RL} & \footnotesize{M} & \footnotesize{U-f}
\\

& \multicolumn{5}{c}{X guides GPT3.5}
\\
\hline

Gen & \footnotesize{23.50} & \footnotesize{8.05} & \footnotesize{21.69} & \footnotesize{22.58} & \footnotesize{32.83}
\\

Exp &  \footnotesize{23.99} & \footnotesize{8.55} & \footnotesize{22.18} & \footnotesize{23.69} & \footnotesize{32.79}
\\

NoExp &  \footnotesize{25.77} & \footnotesize{7.96} & \footnotesize{23.96} & \footnotesize{22.69} & \footnotesize{33.27}
\\

APO-\scriptsize{GPT3.5} &  \footnotesize{24.22} & \footnotesize{9.17} & \footnotesize{22.45} & \footnotesize{22.82} & \footnotesize{32.53}
\\

APO-\scriptsize{GPT4} &  \footnotesize{27.92} & \footnotesize{11.32} & \footnotesize{26.14} & \footnotesize{25.00} & \footnotesize{36.89}
\\

\hline

& \multicolumn{5}{c}{X guides GPT4}
\\
\hline

Gen & \footnotesize{24.99} & \footnotesize{8.94} & \footnotesize{23.74} & \footnotesize{24.82} & \footnotesize{33.13}
\\

Exp &  \footnotesize{24.06} & \footnotesize{8.43} & \footnotesize{21.74} & \footnotesize{25.12} & \footnotesize{31.84}
\\

NoExp &  \footnotesize{23.87} & \footnotesize{7.56} & \footnotesize{22.21} & \footnotesize{23.32} & \footnotesize{31.88}
\\

APO-\scriptsize{GPT3.5} &  \footnotesize{23.19} & \footnotesize{8.31} & \footnotesize{21.59} & \footnotesize{23.79} & \footnotesize{28.94}
\\

APO-\scriptsize{GPT4} &  \footnotesize{30.00} & \footnotesize{11.14} & \footnotesize{27.86} & \footnotesize{26.35} & \footnotesize{35.27}
\\

\hline
\end{tabular}
}
\vspace{-2mm}
\caption{Performance across different prompting groups for GPT3.5 and GPT4.
`Gen' denotes the baseline generic prompts, `Exp' and `NoExp' represent expert and non-expert human modifications, respectively, while `APO-GPT3.5' and `APO-GPT4' indicate prompts refined through APO.}
\label{table:APO_results}
\vspace{-6mm}
\end{table}


\subsection{Experimental Setup}
We put the details of our dataset in Appendix~\ref{dataset-appendix}.
First, we designed the experiment to use the generic prompt, outlined in Appendix \ref{prompts}, on six different GPT models~\footnote{\texttt{text-ada-001}, \texttt{text-babbage-001}, \texttt{text-curie-001}, \texttt{text-davinci-003}, \texttt{gpt-3.5-turbo-0613}, and \texttt{gpt-4-0613}}.
This objective was to evaluate which variants are the best across most sections, thereby guiding our selection for use in APO.
We then divided our experiments into two sets~\footnote{After we got the different sets' prompts, we then ran \texttt{gpt-3.5-turbo-0613} or \texttt{gpt-4-0613} API with self-consistency and zero-shot settings~\cite{wang2022self}, where temperature=0.3, run numbers=5. We used the default numbers for all other parameters in OpenAI API.}:

\textbf{Set-1: Comparative Analysis of APO and Human Contributions in Clinical Note Generation.}
\label{section:exp_set_1}
This experiment aims to assess how APO, compared with humans, can assist in improving content generation for different sections of clinical notes. Specifically, we introduce a generic prompt and training data for distinct sections. The goal is to aid AI systems, such as GPT3.5 and GPT4, in identifying suitable section prompts that enhance content generation in each section. Our experiment involves four groups of prompters: medical experts~\footnote{One licensed physician}, non-medical experts~\footnote{One has a master's degree, and one has a bachelor's degree. They do not have any medical background.}, GPT3.5 (with APO), and GPT4 (with APO). Each group modifies the generic prompt based on the training data for each section. We then compare the effectiveness of these modified prompts in assisting AI to generate summaries for different sections, using the results of the generic prompt as a baseline.

\textbf{Set-2: Enhancing AI-Generated Clinical Content through Humans Prompt Modification Post-APO.}
\label{section:exp_set_2}
In this set of experiments, we take the results of GPT3.5 (with APO) and GPT4 (with APO) as new baselines and invite medical experts and non-medical experts to further modify the prompts based on their knowledge and preferences. This approach examines how human intervention, post-APO implementation, affects the quality of AI-generated content in various clinical note sections. We analyze the effectiveness of these modifications by comparing them against the baseline established by APO-modified prompts, focusing on the nuances introduced by the domain-specific knowledge and preferences of the two human groups.

\begin{table}
\centering
\scalebox{0.7}{
\begin{tabular}{c|ccccc}
\hline
\multirow{2}{*}{Mentor} & \footnotesize{R1} & \footnotesize{R2} & \footnotesize{RL} & \footnotesize{M} & \footnotesize{U-f}
\\

 & \multicolumn{5}{c}{X guides GPT3.5}
\\
\hline

APO-\scriptsize{GPT4} &  \footnotesize{27.92} & \footnotesize{11.32} & \footnotesize{26.14} & \footnotesize{25.00} & \footnotesize{36.89}
\\

Exp-\scriptsize{APO} &  \footnotesize{26.89} & \footnotesize{10.82} & \footnotesize{25.39} & \footnotesize{25.46} & \footnotesize{36.62}
\\

NoExp-\scriptsize{APO} &  \footnotesize{26.71} & \footnotesize{9.07} & \footnotesize{24.89} & \footnotesize{21.68} & \footnotesize{33.44}
\\

\hline

& \multicolumn{5}{c}{X guides GPT4}
\\
\hline

APO-\scriptsize{GPT4} &  \footnotesize{30.00} & \footnotesize{11.14} & \footnotesize{27.86} & \footnotesize{26.35} & \footnotesize{35.27}
\\

Exp-\scriptsize{APO} &  \footnotesize{28.83} & \footnotesize{10.70} & \footnotesize{27.20} & \footnotesize{26.48} & \footnotesize{35.57}
\\

NoExp-\scriptsize{APO} &  \footnotesize{28.28} & \footnotesize{9.78} & \footnotesize{26.60} & \footnotesize{24.25} & \footnotesize{32.68}
\\

\hline
\end{tabular}
}
\vspace{-2mm}
\caption{Comparative effectiveness of post-APO-GPT4 human prompt modifications. This table shows the results of human intervention after APO-GPT4 prompts, where `Exp-APO' and `NoExp-APO' denote the post-APO-GPT4 modifications by experts and non-experts. } 

\label{table:HIL_results}
\vspace{-6mm}
\end{table}

\subsection{Results}
\label{results}
For our initial experiment, the findings indicate that GPT-4 and GPT3.5 emerged as the most effective variants, in descending order of performance, as detailed in Appendix \ref{variants}.
As a result, they were used for our proposed algorithm.

\textbf{Set-1: Comparative Analysis of APO and Human Contributions in Clinical Note Generation.}
Upon examining the `X guides GPT3.5' results from Table~\ref{table:APO_results}~\footnote{The details can be found in Appendix Table \ref{table:diff_section_performance}}, we observed that expert and non-expert modifications resulted in slight improvements compared to the generic (baseline) results. 
However, according to the ROUGE and METEOR scores, `expert guides GPT3.5' did not yield better outcomes than `non-expert guides GPT3.5'; non-experts led regarding factuality (UMLS-f1) scores. 
The performance of APO-GPT3.5 did not significantly differ from the baseline, whereas APO-GPT4 markedly surpassed all other methods.
Compared to human modifications, APO-GPT4 enhanced summary quality, a feat APO-GPT3.5 did not achieve. 
For the same  Table~\ref{table:APO_results} `X guides GPT3.5' experiment, the results indicated that prompts modified by experts, non-experts, and APO-GPT3.5 all fell short of the generic prompt across various sections, with expert modifications slightly outperforming non-experts, and both human groups surpassing APO-GPT3.5, especially in terms of factuality score. 
Consistent with the `X guides GPT3.5' findings, APO-GPT4 again significantly elevated the scores across the board. 
Finally, the results in the Appendix Table~\ref{table:diff_section_performance} show the helpful effect of APO-GPT4 on problem \textcolor{red}{(2)} and \textcolor{red}{(3)} in Figure~\ref{fig:mentor_guide_mentee}.
These results further demonstrate GPT4's emergent abilities in self-critique~\cite{madaan2023self}, self-feedback~\cite{huang2022large}, and self-explanation~\cite{zhao2023self}.

\textbf{Set-2: Enhancing AI-Generated Clinical Content through Humans Prompt Modification Post-APO.}
In this experiment, we continued to explore the outcomes of the human-in-the-loop paradigm on top of APO. From the previous experiments in Table~\ref{table:APO_results}, it was evident that APO-GPT4 significantly boosted the summary quality, raising the lower bound of AI performance on this task and providing a new baseline for users to engage in further prompt engineering. 
We refer to the process of experts post-editing APO-refined APO-GPT4 prompts as `Exp-APO' and the analogous post-editing by non-experts as `NoExp-APO'.
We compared Exp-APO and NoExp-APO modifications, with the term `APO' now exclusively referring to the results achieved by APO-GPT4. 
In Table \ref{table:HIL_results}, we found that for both `X guides GPT3.5' and `X guides GPT4', Exp-APO modifications did not significantly differ from APO-GPT4 in terms of ROUGE, METEOR, and UMLS-f1 scores, whereas NoExp-APO modifications notably degraded summary quality, particularly factuality scores, suggesting a loss of key information or the introduction of hallucinations.

In a detailed \textbf{comparison between Exp-APO and APO-GPT4}, we curated a human evaluation dataset from 100 randomly selected instances within the evaluation set. 
This allowed experts who contributed to Exp-APO to assess and provide feedback on their preference for summaries generated from their revised prompts compared to those produced by the original APO-GPT4 prompts.
The outcome showed a preference distribution where 75\% favored Exp-APO, 3\% indicated no preference, and 22\% preferred APO-GPT4. 
These results show that while factuality scores remained closely comparable, there was a slight decrease in ROUGE scores for Exp-APO, yet the expert preference was markedly in favor of Exp-APO. 
This can be attributed to how APO tends to enforce certain structural elements within prompts, such as explicitly stating `None' in the absence of information. 
Experts tended to remove such repetitive formulations, which, although potentially reducing the strict adherence to format and the ROUGE score, did not impact the factuality score. 
Moreover, experts' preferences are less influenced by rigid formatting and more by their own knowledge and experience. 
These expert insights, incorporated through the human-in-the-loop approach, may have introduced a degree of personalization to the prompts, aligning the AI-generated content more closely with human evaluative criteria and contributing to the overall preference for Exp-APO.
This suggests that while expert post-editing prompts may not markedly enhance the quality of APO-GPT4 summaries, they align more closely with user preferences, offering a more personalized result without sacrificing summary quality.

\section{Conclusion}
Our investigation has demonstrated the profound impact of prompt engineering on the effectiveness of LLMs, specifically in clinical note generation. 
Implementing our APO framework has notably advanced the standardization of prompt quality, particularly with GPT4, which has shown superior performance in generating clinical notes. 
Incorporating a human-in-the-loop approach further validated the importance of expert involvement, indicating a clear preference for expert-modified prompts, suggesting that personalized tweaks to APO-generated prompts yield user-preferred outcomes without compromising the content's integrity.

\section{Limitations}
Our research, while insightful, acknowledges several limitations. The task-specific nature of our findings implies that even if prompts perform well within our dataset, this does not guarantee similar success in real-world, complex scenarios. The MTS-Dialog dataset's limitations also pose challenges; many sections had insufficient data, leading to exclusion and a lack of comprehensive coverage. Even after preprocessing and filtering, data imbalance remains a concern. Moreover, our evaluation metrics—ROUGE, METEOR, and UMLS-f1—may not fully encapsulate the qualitative subtleties of clinical note generation, potentially overlooking nuances apparent to human experts. The number of human mentors involved was constrained by time and financial resources, possibly introducing bias into the results.

Recent advancements in APO have seen the development of more sophisticated algorithms aimed at enhancing efficacy and stability~\cite{fernando2023promptbreeder, wang2023promptagent, dong2023pace, li2023automatic, sun2023autohint,opsahl2024optimizing}; however, these were not compared in our study. Additionally, our approach to prompting with APO and human experts primarily focused on general quality without targeting specific aspects such as hallucination~\cite{huang2023survey}. Tailoring the APO algorithm to improve particular model performances (e.g., factuality) could yield more targeted enhancements. The integration of external resources, like databases, information retrieval systems, or writing assistant tools, could also provide additional information to aid AI in making more accurate suggestions during the forward pass and refinements during the backward pass, overcoming some of the AI's knowledge limitations~\cite{petroni2019language, sung2021can, yao2022extracting, yao2022context, singhal2022large}.

Moving forward, we plan to delve deeper into the nuances of prompt engineering, exploring the boundaries of personalization and the potential for even more sophisticated AI-human collaboration models. We aim to expand the diversity of expert input and examine the impact of such variations on the overall system performance. Furthermore, future work will also investigate the scalability of our approach to other domains within NLP, testing the generalizability and robustness of the APO framework. In addition, we are also interested in the emergent ability of GPT4 that can perform APO for other AI and itself well, and we plan to distill this ability into trainable LLMs, such as the LLaMA family~\cite{touvron2023llama1, touvron2023llama2}, by creating a batch of synthetic instruction learning data~\cite{wang2022self, tran2023bioinstruct}.

\section{Ethics Statement}
In conducting this research, we have adhered to ethical guidelines, ensuring that all patient data used in the dataset was anonymized and used strictly for research purposes. We have also considered the potential implications of our work on clinical practice, emphasizing the enhancement of AI tools as assistive rather than replacement technologies to support medical professionals. As we progress, we remain committed to upholding these ethical standards and continuously assessing the societal impacts of our research.

\bibliography{anthology,custom}

\newpage

\appendix

\section{Appendix}
\label{sec:appendix}

\subsection{SOAP Structure}
\label{SOAP-Structure-appendix}
The SOAP (Subjective, Objective, Assessment, and Plan) structure is commonly used by providers \cite{podder2021soap}.

\begin{enumerate}[topsep=0.5pt,itemsep=0.2ex,partopsep=0.2ex,parsep=.20ex, label=$\ast$]
    \item The Chief Complaint section is a brief description of a patient’s conditions and the reasons for the visit.
    \item The Subjective section is a detailed report of the patient’s current conditions, such as source, onset, and duration of symptoms, mainly based on the patient’s self-report. This section usually includes a history of present illness and symptoms, current medications, and allergies.
    \item The Objective section documents the results of physical exam findings, laboratory data, vital signs, and descriptions of imaging results.
    \item The Assessment section typically contains medical diagnoses and reasons that lead to medical diagnoses. The assessment is typically based on the content of the chief complaint and the subjective and objective sections.
    \item The Plan section addresses treatment plans based on the assessment.
\end{enumerate}

\subsection{Human Annotation Guideline}

\begin{table}[H]
\centering
\scalebox{1}{
\begin{tabular}{l|c}
\hline

SOAP sections & \# Data
\\

\hline
ASSESSMENT & 33
\\

PLAN & 9
\\

EDCOURSE & 6
\\

DISPOSITION & 12
\\

PASTSURGICAL & 66
\\

PASTMEDICALHX &  117
\\

ROS & 66
\\

GENHX &  297
\\

ALLERGY &  59
\\

MEDICATIONS & 55
\\

FAM SOCHX &  368
\\

DIAGNOSIS & 15
\\

CC &  75
\\

EXAM &  19
\\

\hline
Overall &  1197
\\

\hline
\end{tabular}
}
\caption{The data distribution across sections in our evaluation dataset.} 

\label{table:data_size}
\vspace{-4mm}
\end{table}

\begin{figure*}
\centering
\includegraphics[width=\textwidth]{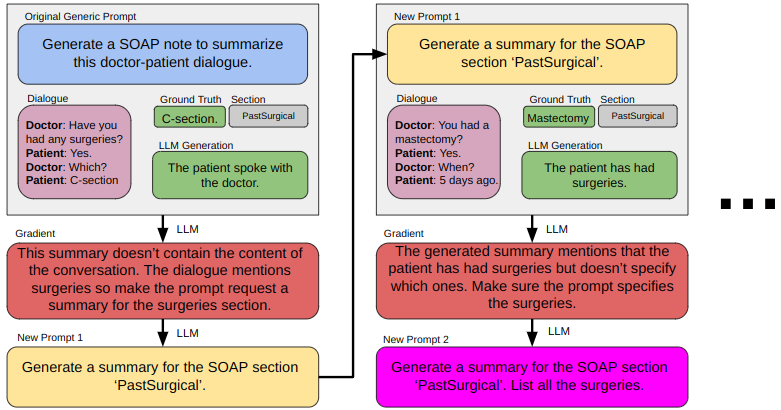}
\caption{Overview and example of a \textbf{correct} APO on clinical note generation. While training on a batch, all the data instances start from the updated prompt based on suggestions from its immediate prior data instance.}
\label{fig:flow}
\vspace{-4mm}
\end{figure*}

\begin{figure*}
\centering
\includegraphics[width=\textwidth]{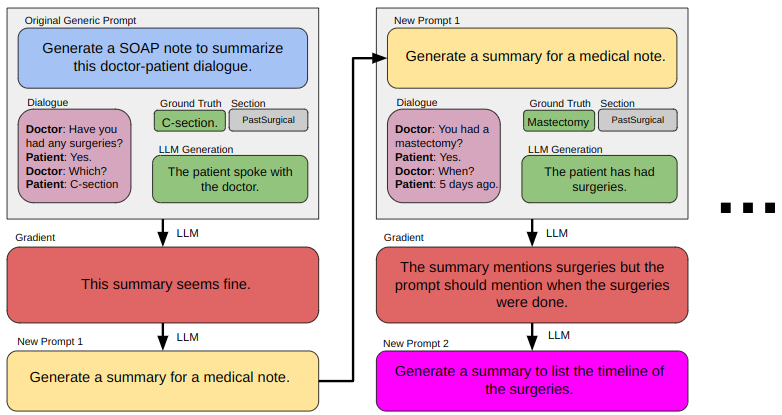}
\caption{Overview and example of an \textbf{incorrect} APO on clinical note generation.}
\label{fig:flow2}
\vspace{-4mm}
\end{figure*}

\begin{table*}
    \centering
    \begin{tabularx}{\textwidth}{p{2.5cm}p{5cm}p{7.2cm}}
    \hline
    \footnotesize{Section} & \footnotesize{Subsection} & \footnotesize{Definition}
    \\
    \hline
    \footnotesize{Subjective} & &
    \\

    & \footnotesize{Chief Complaint} 
    & \footnotesize{Patient’s primary motivation for the visit and type of visit}
    \\
    & \footnotesize{Review of Systems}		
    & \footnotesize{Patient’s report of system-related health and symptoms}
    \\
    & \footnotesize{Past Medical History}
    & \footnotesize{Patient’s reported diagnoses/conditions (when and what,  excluding laboratory and imaging results and surgeries)}
    \\
    & \footnotesize{Past Surgical History}
    & \footnotesize{Patient’s reported prior surgeries (what, when, where)}
    \\
    & \footnotesize{Family Medical History}
    & \footnotesize{Conditions affecting patient’s close genetic relatives}
    \\
    & \footnotesize{Social History}
    & \footnotesize{Patient’s alcohol, tobacco, and drug-related behaviors}
    \\
    & \footnotesize{Medications}
    & \footnotesize{Patient’s list of medications (not prescribed during visit)}
    \\
    & \footnotesize{Allergies}
    & \footnotesize{Patient’s list of allergies (primarily medicinal)}
    \\
    & \footnotesize{Miscellaneous}
    & \footnotesize{Patient’s clinically relevant social and other circumstances}
    \\
    \hline
    \footnotesize{Objective} & &
    \\
    & \footnotesize{Immunizations}
    & \footnotesize{Vaccination record (not frequently discussed)}
    \\
    & \footnotesize{Laboratory and Imaging Results}
    & \footnotesize{Clinician’s discussion of laboratory/imaging results}
    \\
    \hline

    \footnotesize{Assessment} & &
    \\
    & \footnotesize{Assessment}
    & \footnotesize{Synthesis of the reason for the visit and pertinent diagnosis}
    \\
    \hline
    \footnotesize{Plan} & &
    \\
    & \footnotesize{Diagnostics \& Appointments}
    & \footnotesize{Plan for future tests, appointments, or surgeries}
    \\
    & \footnotesize{Prescriptions \& Therapeutics}
    & \footnotesize{Plan for medications and therapeutics}
    \\

    \hline
    \end{tabularx}
    \caption{Details of the SOAP structure used in our CC and CCUser datasets.} 
    \label{table:SOAP-structure-appendix}
\end{table*}

\begin{table*}
\centering
\scalebox{0.86}{
\begin{tabular}{cIc|ccc|cIc|ccc|c}
\hline

\hline
& \multicolumn{5}{cI}{X guides GPT3.5} & \multicolumn{5}{c}{X guides GPT4}
\\
\hline

SOAP sections & GEN & \footnotesize{Human1} & \footnotesize{Human2} & \footnotesize{Human3} & APO & GEN & \footnotesize{Human1} & \footnotesize{Human2} & \footnotesize{Human3} & APO
\\

\hline
\scriptsize{ASSESSMENT} & 18.77 & \colorbox{orange3}{+1.27} & \colorbox{blue1}{-0.16} & \colorbox{orange1}{+0.09} & \colorbox{orange1}{+0.37} & 17.44	& \colorbox{blue3}{-1.67}	& \colorbox{blue5}{-5.33}	& \colorbox{blue1}{-0.97}	& \colorbox{blue3}{-1.7} 
\\

\scriptsize{PLAN} & 17.64 & \colorbox{orange5}{+5.05} & \colorbox{orange5}{+5.42} & \colorbox{orange5}{+5.12} & \colorbox{orange5}{+5.59} & 22.01	& \colorbox{orange1}{+0.17}	& \colorbox{blue3}{-1.59}	& \colorbox{orange1}{+0.21}	& \colorbox{orange5}{+4.12}
\\

\scriptsize{EDCOURSE} & 31.16 & \colorbox{blue2}{-2.87} & \colorbox{orange1}{+0.3} & \colorbox{orange3}{+3.16} & \colorbox{orange3}{+3.34} & 38.2  & \colorbox{blue3}{-3.51}	& \colorbox{blue3}{-2.66}	& \colorbox{blue3}{-2.87}	& \colorbox{blue3}{-2.68}
\\

\scriptsize{DISPOSITION} & 16.00 & \colorbox{orange3}{+3.48} & \colorbox{blue2}{-1.71} & \colorbox{blue1}{-0.07} & \colorbox{orange5}{+4.92} & 17.14	& \colorbox{orange3}{+2.88}	& \colorbox{orange5}{+4.86}	& \colorbox{blue3}{-1.19}	& \colorbox{blue3}{-1.07}
\\

\scriptsize{PASTSURGICAL} & 22.42 & \colorbox{orange3}{+1.28} & \colorbox{orange3}{+4.89} & \colorbox{orange7}{+11.53} & \colorbox{orange5}{+4.36} & 23.06	& \colorbox{blue3}{-2.05}	& \colorbox{blue1}{-0.86}	& \colorbox{blue1}{-0.41}	& \colorbox{orange3}{+1.9}
\\

\scriptsize{PASTMEDICALHX} &  23.62 & \colorbox{orange1}{+0.64} & \colorbox{orange1}{+0.61} & \colorbox{orange3}{+2.79} & \colorbox{orange3}{+2.78} & 25.19	& \colorbox{orange1}{+0.07}	& \colorbox{blue1}{-0.19}	& \colorbox{orange1}{+0.1}	& \colorbox{orange1}{+0.4}
\\

\scriptsize{ROS} &  29.01 & \colorbox{orange1}{+0.58} & \colorbox{blue1}{-0.04} & \colorbox{orange1}{+0.14} & \colorbox{orange1}{+0.61} & 29.79	& \colorbox{orange1}{+0.06}	& \colorbox{blue5}{-6.86}	& \colorbox{blue3}{-2.77}	& \colorbox{blue3}{-1.45}
\\

\scriptsize{GENHX} &  40.21 & \colorbox{orange3}{+1.66} & \colorbox{blue3}{-2.53} & \colorbox{orange3}{+2.16} & \colorbox{orange1}{+0.74} & 43.27	& \colorbox{orange1}{+0.1}	& \colorbox{blue5}{-4.93}	& \colorbox{blue3}{-2.44}	& \colorbox{blue3}{-3.95}
\\

\scriptsize{ALLERGY} &  21.48 & \colorbox{blue3}{-1.89} & \colorbox{blue1}{-0.94} & \colorbox{orange7}{+8.93} & \colorbox{OrangeRed}{+24.58} & 28.29	& \colorbox{blue1}{-0.8}	& \colorbox{orange1}{+0.96}	& \colorbox{orange1}{+0.26}	& \colorbox{orange9}{+14.2}
\\

\scriptsize{MEDICATIONS} &  20.14 & \colorbox{blue3}{-1.15} & \colorbox{orange1}{+0.82} & \colorbox{OrangeRed}{+27.44} & \colorbox{orange7}{+6.78} & 19.81	& \colorbox{blue7}{-7.59}	& \colorbox{blue3}{-2.07}	& \colorbox{orange5}{+4.87}	& \colorbox{OrangeRed}{+24.72}
\\

\scriptsize{FAM SOCHX} &  31.63 & \colorbox{blue1}{-0.64} & \colorbox{blue3}{-1.66} & \colorbox{blue5}{-3.92} & \colorbox{blue3}{-1.3} & 30.71	& \colorbox{blue1}{-0.71}	& \colorbox{blue1}{-0.82}	& \colorbox{blue7}{-7.91}	& \colorbox{blue1}{-0.19}
\\

\scriptsize{DIAGNOSIS} &  17.81 & \colorbox{blue3}{-1.54} & \colorbox{orange1}{+0.93} & \colorbox{orange1}{+0.35} & \colorbox{blue1}{-0.13} & 16.4	& \colorbox{blue3}{-2.93}	& \colorbox{orange5}{+4.35}	& \colorbox{orange1}{+0.59}	& \colorbox{orange7}{+8.87}
\\

\scriptsize{CC} &  16.09 & \colorbox{blue1}{-0.64} & \colorbox{blue1}{-0.54} & \colorbox{blue1}{-0.68} & \colorbox{orange5}{+3.99} & 15.17	& \colorbox{orange3}{+1.85}	& \colorbox{orange3}{+2.92}	& \colorbox{orange3}{+3.7}	& \colorbox{OrangeRed}{+22.12}
\\

\scriptsize{EXAM} &  23.30 & \colorbox{orange3}{+1.4} & \colorbox{orange3}{+2.71} & \colorbox{blue3}{-1.86} & \colorbox{orange5}{+4.94} & 23.47	& \colorbox{orange3}{+1.04}	& \colorbox{blue3}{-1.92}	& \colorbox{blue9}{-10.2}	& \colorbox{orange5}{+4.85}
\\

\hline
Overall &  23.50 & \colorbox{orange1}{+0.49} & \colorbox{orange1}{+0.59} & \colorbox{orange3}{+3.96} & \colorbox{orange5}{+4.42} & 24.99	& \colorbox{blue1}{-0.93}	& \colorbox{blue1}{-0.88}	& \colorbox{blue3}{-1.36}	& \colorbox{orange5}{+5.01}
\\

\hline
\end{tabular}
}
\caption{Different sections' performance across different prompting groups for GPT3.5 and GPT4. 
This is the ROUGE1 full table for Figure \ref{fig:mentor_guide_mentee}, and Table \ref{table:APO_results}.`Gen' denotes the baseline generic prompts. `Human1', `Human2', and `Human3' denote different humans's prompting engineering results over the generic prompt. 
The number here is the increment compared to GEN after prompting. Orange/red represents an increase, blue represents a decrease. The darker the color, the greater the increment.
} 

\label{table:diff_section_performance}
\vspace{-4mm}
\end{table*}

\begin{table*}
\centering
\scalebox{0.8}{
\begin{tabular}{cIcIccc|c|cIccc}
\hline

\hline
\multirow{2}{4em}{ROUGE1} & & \multicolumn{5}{cI}{X guides GPT3.5} & \multicolumn{3}{c}{\small{X post-edit APO-guides-GPT3.5}}
\\

& GEN & \footnotesize{Human1} & \footnotesize{Human2} & \footnotesize{Human3} & GPT3.5 & GPT4 & \footnotesize{Human1} & \footnotesize{Human2} & \footnotesize{Human3}
\\

\hline
ASSESSMENT & 18.77 & 20.04 & 18.61 & 18.86 & 19.39 & 19.14 &	18.99 &	19.52 & 19.13
\\
PLAN & 17.46 & 22.69 &	23.06 & 22.76 &	23.45 &	23.23 &	22.42 &	20.69 &	23.1
\\
EDCOURSE&	31.16 &	28.29 &	31.46 &	34.32 &	35.15 &	34.5 & 34.84 & 26.61 &	32.83
\\
DISPOSITION &	16 & 19.48 & 14.29 & 15.93 & 19.34 & 20.92 &	19.18 &	14.58 &	16.67
\\
PASTSURGICAL&	22.42 &	23.7 & 27.31 & 33.95 & 25.93 & 26.78 &	26.21 &	30.8 & 32.94
\\
PASTMEDICALHX&	23.62 &	24.26 &	24.23 &	26.41 &	19.85 &	26.4 &	25.78 &	22.06 &	26.16
\\
ROS&	29.01 & 29.59 &	28.97 &	29.15 &	14.31 &	29.62 &	25.78 &	24.59 &	30.34
\\
GENHX&	40.21 &	41.87 &	37.68 &	42.37 &	42.76 &	40.95 &	40.83 &	39.14 &	42.01
\\
ALLERGY& 21.48 & 19.59 & 20.54 & 30.41 & 34.66 & 46.06 & 44.86	& 45.27 & 31.76
\\
MEDICATIONS&	20.14 &	18.99 &	20.96 &	47.58 &	17.25 &	26.92 &	27.15 &	20.27 &	48.78
\\
FAM SOCHX&	31.63 &	30.99 &	29.97 &	27.71 &	30.96 &	30.33 &	30.13 & 29.79 & 30.49
\\
DIAGNOSIS&	17.81 &	16.27 &	18.74 &	18.16 &	15.22 &	17.68 &	17.57 & 16.33 & 17.27
\\
CC& 16.09 &	15.45 &	15.55 &	15.41 &	17.61 &	20.08 &	18.05 &	15.02 & 21.24
\\
EXAM& 23.3	& 24.7	& 26.01	& 21.44 & 23.29 & 28.24 & 24.67 &	26.15 &	24.51
\\
\hline

Overall	&23.5 &	23.99 &	24.09 &	27.46 &	24.22 &	27.92 &	26.89 &	25.06 &	28.37
\\
\hline

\multirow{2}{4em}{ROUGE2} & & \multicolumn{5}{cI}{X guides GPT3.5} & \multicolumn{3}{c}{\small{X post-edit APO-guides-GPT3.5}}
\\

& GEN & \footnotesize{Human1} & \footnotesize{Human2} & \footnotesize{Human3} & GPT3.5 & GPT4 & \footnotesize{Human1} & \footnotesize{Human2} & \footnotesize{Human3}
\\

\hline
ASSESSMENT & 5.94	& 6.45 & 7.05 & 5.52 &	6.79 &	6.52 &	5.75 & 6.69 &	6.21
\\
PLAN & 5.76 & 8.11 & 7.78 &	9.3	& 8.99 & 7.45 &	10.26 &	8.1 &	7.75
\\
EDCOURSE&	12.11 &	12 & 11.46 & 14.15 & 12.89 & 13.35 &	13.36 & 11.04 & 12.09
\\
DISPOSITION& 3.46 &	7.46 &	2.84 &	4.5	& 7.53 & 13.86 &	8.02 & 3.71 & 1.75
\\
PASTSURGICAL&	8.63 &	10.12 &	12.18 &	9.34 &	10.18 &	11.59 &	10.83 &	8.98 &	9.65
\\
PASTMEDICALHX&	8.7 & 8.19 & 8.49 & 9.86 & 6.1 & 9.73 &	9.09 &	6.92 & 10.08
\\
ROS&	8.24 &	8.54 &	8.21 &	8.34 &	3.93 &	8.71 &	8.88 &	6.86 &	8.86
\\
GENHX&	14.11 &	14.86 &	12.28 &	15.21 &	15.73 &	14.37 &	14.33 &	13.62 &	14.94
\\
ALLERGY	& 8.41 & 8.55 &	7.06 & 2.74 & 22.34 & 29.83 & 30.2 &	30.55 &	3.11
\\
MEDICATIONS&	7.51 &	6.46 &	7.37 &	4.87 &	5.24 &	9.3 &	9.74 &	6.85 &	11.55
\\
FAM SOCHX&	13.26 &	12.85 &	11.8 & 10.19 &	12.74 &	11.83 &	11.61 & 11.97 & 11.85
\\
DIAGNOSIS&	5.37 &	5.6 & 5.63 & 5.48 &	4.33 &	6.04 &	6.04 &	4.75 & 5.51
\\
CC	& 4.49	& 3.68 & 3.81 &	3.59 & 5.1	& 6.87 & 5.14 &	4.37 & 8.23
\\
EXAM	& 6.71 & 6.86 &	8.06 & 5.86 & 6.48 & 9.11 &	8.27 &	8.75 & 9.26
\\
\hline

Overall	& 8.05 & 8.55 &	8.14 & 7.78 & 9.17 & 11.32 & 10.82 & 9.51 & 8.63
\\

\hline

\multirow{2}{4em}{ROUGEL} & & \multicolumn{5}{cI}{X guides GPT3.5} & \multicolumn{3}{c}{\small{X post-edit APO-guides-GPT3.5}}
\\

& GEN & \footnotesize{Human1} & \footnotesize{Human2} & \footnotesize{Human3} & GPT3.5 & GPT4 & \footnotesize{Human1} & \footnotesize{Human2} & \footnotesize{Human3}
\\

\hline
ASSESSMENT & 17.24	& 18.31 & 17.65 & 16.95 & 17.73 & 17.62 &	17.47 &	17.51 &	17.76
\\
PLAN & 15.73 & 19.53 & 19.97 & 20.58 & 20.84 & 20.5	& 20.48 &	18.01 &	20.55
\\
EDCOURSE&	28.17 &	27.02 &	29.86 &	31.84 &	33.15 &	33.17 &	33.21 & 25.14 & 29.95
\\
DISPOSITION& 16 & 19.27 & 14.05 & 15.93 & 19.11 & 20.92 &	19.18 & 14.58 & 16.67
\\
PASTSURGICAL&	20.51 &	21.6 & 25.35 & 32.59 & 24.11 & 24.9 &	24.24 &	28.79 &	31.08
\\
PASTMEDICALHX&	21.27 &	21.86 &	21.74 &	23.46 &	18.39 &	24.32 &	23.56 &	20.25 &	24.03
\\
ROS&	25.36 &	26.37 &	25.54 &	25.83 &	12.86 &	26.35 &	26.59 &	22.4 & 27.02
\\
GENHX&	37.4 & 38.94 & 34.88 & 39.4	& 39.68 & 38 & 37.98 &	36.38 & 39.02
\\
ALLERGY	& 20.79 & 19.2 & 19.92 & 30.2	& 34.42 & 45.9 &	44.62 & 44.91 & 31.65
\\
MEDICATIONS&	19.18 &	18.19 &	20.05 &	47.37 &	16.18 &	25.49 &	25.74 &	19.37 &	47.83
\\
FAM SOCHX&	29.6 &	29.16 &	28.02 &	25.69 &	29.03 &	28.16 &	27.95 & 27.98 & 28.45
\\
DIAGNOSIS&	15.2 &	13.31 &	15.88 &	14.81 &	12.02 &	14.45 &	14.34 &	13.1 & 	13.72
\\
CC	& 14.89 & 14.42 & 14.42 & 14.39 & 16.55 & 18.67 & 16.88 &	14.12 &	19.73
\\
EXAM & 22.32 &	23.44 &	24.6 &	20.09 &	20.23 &	27.51 &	23.22 &	23.76 &	23.35
\\
\hline

Overall	& 21.69 &	22.18 &	22.28 &	25.65 &	22.45 &	26.14 &	25.39 &	23.31 &	26.48
\\

\hline

\end{tabular}
}
\caption{Different sections' performance across different prompting groups for GPT3.5. 
This is the ROUGE1, 2, L full table for Table \ref{table:APO_results}, and Table \ref{table:HIL_results}
.} 

\label{table:full_results1}
\end{table*}

\begin{table*}
\centering
\scalebox{0.8}{
\begin{tabular}{cIcIccc|c|cIccc}
\hline

\hline
\multirow{2}{4em}{METEOR} & & \multicolumn{5}{cI}{X guides GPT3.5} & \multicolumn{3}{c}{\small{X post-edit APO-guides-GPT3.5}}
\\

& GEN & \footnotesize{Human1} & \footnotesize{Human2} & \footnotesize{Human3} & GPT3.5 & GPT4 & \footnotesize{Human1} & \footnotesize{Human2} & \footnotesize{Human3}
\\

\hline
ASSESSMENT & 20.99 & 22.57	& 24.6 & 20.95 & 22.41 & 22.95 &	19.61 &	21.77 &	22.83
\\
PLAN & 17.31 &	23.09 &	22.57 &	25.03 &	20.57 &	19.53 &	23.54 &	19.98 &	21.8
\\
EDCOURSE&	20.57 &	19.48 &	22.93 &	23.32 &	23.52 &	24.08 &	24.65 & 19.43 & 23.55
\\
DISPOSITION& 23.52 & 28.33 & 23.23 & 28.82 & 27.14 & 12.32 &	25.34 &	20.61 &	3.89
\\
PASTSURGICAL&	22.54 &	24.76 &	26.53 &	17.19 &	22.89 &	29.07 &	27.1 & 19.36 &	3.89
\\
PASTMEDICALHX&	21.25 &	22.04 &	22.03 &	23.15 &	19.6 & 22.84 &	21.98 &	20.15 &	23.26
\\
ROS&	21.63 &	22.17 &	21.37 &	21 & 9.32 &	22.73 &	23.08 &	16.54 & 22.84
\\
GENHX&	26.39 &	28.68 &	23.91 &	28.96 &	29.33 & 27.6 & 27.58 &	26.77 &	28.69
\\
ALLERGY	& 23.04 & 23.33 & 21.99 & 10.93 & 31.49 & 42.76 &	42.63 & 39.36	& 9.61
\\
MEDICATIONS& 22.09 & 22.08 & 23.01 & 10.34 & 15.57 & 22.01 & 22.15 & 21.47 & 18.84
\\
FAM SOCHX&	28.75 &	29.28 &	26.88 &	25.39 &	28.49 &	26.33 &	26.16 & 28.45 & 26.54
\\
DIAGNOSIS&	22.99 &	22.37 &	27.53 &	27.24 &	20.91 &	25.08 &	24.97 & 26.11 & 23.79
\\
CC	& 21.06 & 19.48 & 19.29 & 21.21 & 24.45 & 24.9 & 22.33 &	20.59 &	24.04
\\
EXAM & 24.04 &	24.1 & 25.23 & 20.73 & 23.88 & 27.82 & 25.28 &	26.44 &	26.47
\\
\hline

Overall	& 22.58 & 23.69 & 23.65 & 21.73 & 22.82 & 25 & 25.46 &	23.36 &	20
\\

\hline

\multirow{2}{5em}{UMLS-F1} & & \multicolumn{5}{cI}{X guides GPT3.5} & \multicolumn{3}{c}{\small{X post-edit APO-guides-GPT3.5}}
\\

& GEN & \footnotesize{Human1} & \footnotesize{Human2} & \footnotesize{Human3} & GPT3.5 & GPT4 & \footnotesize{Human1} & \footnotesize{Human2} & \footnotesize{Human3}
\\

\hline
ASSESSMENT & 29.43 & 30.78 & 26.29 & 26.28 & 26.87 & 28.78 &	32.66 &	27.29 &	29.48
\\
PLAN & 28.94 & 32.57 & 32.54 & 30.81 & 35.08 &	33.98 &	31.86 &	32.56 &	35.29
\\
EDCOURSE&	29.83 &	31.7 & 36.98 &	32.04 &	38.5 & 37.25 &	38.31 & 31.37 & 35.62
\\
DISPOSITION& 33.43 & 33.34 & 37.47 & 38.32 & 38.62 & 29.72 &	27.23 &	26.4 & 36.11
\\
PASTSURGICAL&	29.66 & 29.02 &	32.75 &	34.39 &	29.9 & 35.18 &	35.29 &	32.7 &	31.27
\\
PASTMEDICALHX&	33.93 &	34.3 & 34.2	& 36.26 & 28.99 & 37.22 &	37.01 &	32.84 &	37.35
\\
ROS&	36.71 &	37.84 &	34.66 &	34.86 &	14.36 &	37.95 &	38.13 &	25.7 &	36.75
\\
GENHX&	43.97 &	45.42 & 40.66 &	45.97 &	45.72 &	44.91 &	44.67 &	41.66 &	45.75
\\
ALLERGY	& 27.4	& 18.66 &	25.29 &	12.75 &	39.51 &	46.57 &	46.59 & 47.14 & 12.85
\\
MEDICATIONS& 39.88 & 38.07 & 39.84 & 49.73 & 33.08 & 45.43 &	45.99 &	38.47 &	41.45
\\
FAM SOCHX&	34.48 &	35.23 &	33.12 &	30.39 &	33.81 &	33.88 &	33.65 & 32.9	 &33.59
\\
DIAGNOSIS&	36.11 &	37.73 & 34.5 & 37.83 &	35.35 &	40	& 38.73	& 30.7 &	41.17
\\
CC	& 28.49	& 27.95 &	29 &	25.2 &	31.57 &	33.73 &	31.76 &	27.35 &	36.17
\\
EXAM & 27.4	& 26.5 & 31.29 & 28.22 & 24.13 & 31.84 & 30.86&	24.99 & 31.62
\\
\hline

Overall	& 32.83 &	32.79 &	33.47 &	33.07 &	32.53 &	36.89 &	36.62 &	32.29 &	34.6
\\

\hline
\end{tabular}
}
\caption{Different sections' performance across different prompting groups for GPT3.5. 
This is the METEOR and UMLS-F1 full table for Table \ref{table:APO_results}, and Table \ref{table:HIL_results}
.} 

\label{table:full_results2}
\end{table*}

\begin{table*}
\centering
\scalebox{0.8}{
\begin{tabular}{cIcIccc|c|cIccc}
\hline

\hline
\multirow{2}{4em}{ROUGE1} & & \multicolumn{5}{cI}{X guides GPT4} & \multicolumn{3}{c}{\small{X post-edit APO-guides-GPT4}}
\\

& GEN & \footnotesize{Human1} & \footnotesize{Human2} & \footnotesize{Human3} & GPT3.5 & GPT4 & \footnotesize{Human1} & \footnotesize{Human2} & \footnotesize{Human3}
\\

\hline
ASSESSMENT & 17.44 & 15.77 & 12.11 & 16.47 & 17.28 & 15.74	& 16.72	& 15.16	& 15.49
\\
PLAN & 22.01 & 22.18 & 20.42 & 22.22 & 22.88 & 26.13 & 25.9 & 25.9	& 25.86
\\
EDCOURSE&	38.2&	34.69&	35.54&	35.33&	24.91&	35.52&	37.43&	34.98&	34.35
\\
DISPOSITION&	17.14&	20.02&	22.01&	15.95&	11.97&	16.07&	19.31&	15.45&	16.3
\\
PASTSURGICAL&	23.06&	21.04&	22.2&	22.65&	28.12&	24.96	&22.14&	26.9&	33.94
\\
PASTMEDICALHX&	25.19&	25.26&	25&	25.29&	20.37&	25.59&	25.19&	19.58	&24.84
\\
ROS&	29.79&	29.85&	22.93&	27.02& 28.85&	28.34&28.54 &28.91	&28.23
\\
GENHX&	43.27&	43.37&	38.34&40.83	&40.97 &39.32	&39.63&37.7&	40.88
\\
ALLERGY	&28.29	&27.49	&29.25	&28.55	&42.23	&42.49	&42.58	&42.64	&33.57
\\
MEDICATIONS&	19.81	&12.22	&19.54	&24.68	&14.33	&44.53&	44.28	&40.92	&46.36
\\
FAM SOCHX&	30.71&	30	&29.89	&22.8	&25.8	&30.52&	24.22&	24.62&	31.25
\\
DIAGNOSIS&	15.17&	17.02	&18.09	&18.87	&13.76	&37.29	&37.15	&29.14	&21.43
\\
CC	&16.4	&13.47	& 20.75&	16.99&	13.96	&25.27	&16.08&	22.15	&29.11
\\
EXAM	&23.47	&24.51	&21.55	&13.27	&19.27	&28.32	&24.49&	28.16	&18.11
\\
\hline

Overall	&24.99&	24.06	&24.11	&23.63	&23.19	&30	&28.83	&28.01	&28.55
\\
\hline

\multirow{2}{4em}{ROUGE2} & & \multicolumn{5}{cI}{X guides GPT4} & \multicolumn{3}{c}{\small{X post-edit APO-guides-GPT4}}
\\

& GEN & \footnotesize{Human1} & \footnotesize{Human2} & \footnotesize{Human3} & GPT3.5 & GPT4 & \footnotesize{Human1} & \footnotesize{Human2} & \footnotesize{Human3}
\\

\hline
ASSESSMENT & 4.8 & 5.01	& 2.8	&4.88	&5.13	&5.28	&5.36	&4.58&	4.78
\\
PLAN & 9.29	& 9.86 & 8.23 &	9.02 & 9 &12.27	&12.9	&12.82	&12.98
\\
EDCOURSE&	16.04	&13.59	&15.92	&13.5	&8.25 &	14.49	&15.32	& 12.87	& 14.2
\\
DISPOSITION& 3.22	& 5.3	&6.57	&3.47	&1.3	&3.99	&5.4	&3.99	&4.8
\\
PASTSURGICAL&	9.94	&8.43	&9.06	&6.54&	10.16	&11.65	&8.69	&12.41	&11.98
\\
PASTMEDICALHX&	8.48	&8.43	&8.59&	9.19	&6.35	&8.9	&8.72	&6.16	&8.34
\\
ROS&	8.59&	8.86	&6.48	&7.22	&8.5	&8.33&	8.13&	8.16	&8.79
\\
GENHX&	15.96	&15.99	&12.55	&14.1	&14.52	&12.65&	12.88&	12.24&	13.63
\\
ALLERGY	&5.69&	6.09	&5.78	&4.05	&3.22	&9.02	&13.31	&9.58	&6.14
\\
MEDICATIONS&	12.56&	12.59	&13.36&	1.67	&29.29	&29.49&	29.29	&28.62	&3.1
\\
FAM SOCHX&	6.67	&3.65	&6.63	&0.89	&4.24	&8.91	&8.76	&6.78	&9.38
\\
DIAGNOSIS&	12.6	&11.75	&11.63	&8.07	&9.23	&11.85	&8.35	&7.43	&12.48
\\
CC	& 4.16	& 3.34	& 5.78 &	5.62	&3.11	&10.6	&4.56&	8.16	&14.08
\\
EXAM	&7.22	&5.15	&5.68	&4.67	&4.08	&8.52	&8.23&	8.94	&6.66
\\
\hline

Overall	&8.94	&8.43	&8.5	&6.63	&8.31	&11.14	&12.07	&10.19	&9.38
\\

\hline

\multirow{2}{4em}{ROUGEL} & & \multicolumn{5}{cI}{X guides GPT4} & \multicolumn{3}{c}{\small{X post-edit APO-guides-GPT4}}
\\

& GEN & \footnotesize{Human1} & \footnotesize{Human2} & \footnotesize{Human3} & GPT3.5 & GPT4 & \footnotesize{Human1} & \footnotesize{Human2} & \footnotesize{Human3}
\\

\hline
ASSESSMENT & 15.78 &	5.15 &	11.57	&14.64 &	15.39 &	8.52&	15.09 &	13.58 &	13.84
\\
PLAN & 19.46 &	20.12 &	17.44 &	19.16 &	19.8 &	23.06 &	22.84	& 22.84 &	23.16
\\
EDCOURSE&	36.83 &	33.57 &	33.69 & 34.45 &	23.08 &	34.62 &	35.47 &	33.51 &	33.34
\\
DISPOSITION& 16.91 & 19.79 & 21.78 & 15.95 & 11.57	& 16.07 &	19.07 &	15.45 &	16.3
\\
PASTSURGICAL&	21.63 &	19.32 &	20.86 &	21.94 &	26.34 &	23.25 &	20.43 &	25.28 &	32.14
\\
PASTMEDICALHX&	21.63 &	23.03 &	22.81 &	22.56 & 18.74 &	22.96 &	22.81 &	17.64 &	22.15
\\
ROS&	21.63 &	26.86 &	20.97 &	24 & 25.67 & 25.98 & 26.32 &	26.22 &	26.21
\\
GENHX&	40.11 &	40.17 &	35.44 &	37.72 &	37.98 &	36.42 &	36.52 &	34.88 &	37.68
\\
ALLERGY	& 40.11 & 27.13 & 28.9	&28.42 &	41.94 &	42.22 &	42.32 &	42.39 &	33.4
\\
MEDICATIONS&	18.73 &	11.6 & 18.39 &	24.61 &	13.85 &	44.11 &	43.86 &	39.88 &	45.92
\\
FAM SOCHX&	28.54 &	27.87 &	27.81 &	21.11 &	24.12 &	28.32 &	22.54 &	22.9 & 29.12
\\
DIAGNOSIS&	13.9 & 15.64 &	14.64 &	16.58 &	12.94 &	35.18 &	35.94 &	27.49 &	18.75
\\
CC	& 15.3 & 12.31 & 18.62 & 14.55 & 12.6 &	23.24 &	15 & 20.02 &	27.31
\\
EXAM &21.92 & 21.93 & 21.14	& 12.31 & 18.28 & 26.18 & 22.62 &	26.12 &	17.33
\\
\hline

Overall	& 23.74	 &21.74 & 22.43 & 22 & 21.59 & 27.86 &	27.2	&26.3&26.9
\\

\hline

\end{tabular}
}
\caption{Different sections' performance across different prompting groups for GPT4. 
This is the ROUGE1, 2, L full table for Table \ref{table:APO_results}, and Table \ref{table:HIL_results}
.} 

\label{table:full_results3}
\end{table*}

\begin{table*}
\centering
\scalebox{0.8}{
\begin{tabular}{cIcIccc|c|cIccc}
\hline

\hline
\multirow{2}{4em}{METEOR} & & \multicolumn{5}{cI}{X guides GPT4} & \multicolumn{3}{c}{\small{X post-edit APO-guides-GPT4}}
\\

& GEN & \footnotesize{Human1} & \footnotesize{Human2} & \footnotesize{Human3} & GPT3.5 & GPT4 & \footnotesize{Human1} & \footnotesize{Human2} & \footnotesize{Human3}
\\

\hline
ASSESSMENT & 19.69 & 18.77 & 15.05 & 17.67 & 19.1 &	19.06 &	20.28 &	18.04 &	18.81
\\
PLAN & 22.62 &	25.27 &	21.66 &	26.74 &	22.49 &	23.07 &	23.81 &	23.8 & 24.3
\\
EDCOURSE&	26.72 &	26.07 &	26.7 & 28.43 &	18.78 &	25.55 &	26.67 &	25.57 &	26.55
\\
DISPOSITION& 22.81 & 25.35 & 31.92 & 19.23 & 19.34 & 25.65 &	26.24 &	24.78 &	25.03
\\
PASTSURGICAL&	27.59 &	25.68 &	26.87 &	11.21 &	26.28 &	27.67 & 26.84 &	30.59 &	25.24
\\
PASTMEDICALHX&	23.38 &	24.91 &	23.79 &	24.3 & 20.49 &	24.07 &	23.88 &	15.96 &	23.49
\\
ROS&	24.13 &	23.36 &	20.68 &	20.09 &	22.7 & 23.7 & 23.82 &	23.55 &	23.33
\\
GENHX&	30.48 &	30.87 &	29.44 &	28.65 &	30.13 &	29.52 &	29.69 &	29.14 &	30.6
\\
ALLERGY	& 30.48 & 37.42 & 40.43 & 5.86	& 43.96 &	41.56 &	44.3 &	42.55 &	4.32
\\
MEDICATIONS& 22.77 & 16.67 & 40.43 & 2.99 &	20.22 & 19.48 &	19.61 &	21.07 &	17
\\
FAM SOCHX&	29.33 &	30.19 &	29.1 &	21.52 &	26.64 &	29.01 &	25.8 &	20.67 &	29.45
\\
DIAGNOSIS&	22.16 &	26.86 &	26.57 &	30.39 &	22.55 &	32.69 &	35.95 &	34.53 &	32.16
\\
CC	& 22.16 & 16.79 & 23.82 & 23.79 & 18.95 & 23.77 & 20.74 &	24.81 &	19.73
\\
EXAM & 23.24 &	23.57 &	22.8 & 12.85 &	21.52 &	24.22 &	23.08 &	24.05 &	19.99
\\
\hline

Overall	& 24.82 & 25.12 & 27.09 & 19.55 & 23.79 & 26.35 & 26.48 &	25.65 &	22.85
\\

\hline

\multirow{2}{5em}{UMLS-F1} & & \multicolumn{5}{cI}{X guides GPT4} & \multicolumn{3}{c}{\small{X post-edit APO-guides-GPT4}}
\\

& GEN & \footnotesize{Human1} & \footnotesize{Human2} & \footnotesize{Human3} & GPT3.5 & GPT4 & \footnotesize{Human1} & \footnotesize{Human2} & \footnotesize{Human3}
\\

\hline
ASSESSMENT & 32.1 & 25.84 &	19.55 & 3.09 &	27.71 &	26.28 &	30.68 & 26.16 & 26.57
\\
PLAN & 31.91 & 29.73 & 24.87 &	30.55 &	31.22 &	27.15 &	20.28 &	20.28 &	19.99
\\
EDCOURSE&	37.12 &	39.34 &	39.99 &	34.46 &	23.85 &	37.26 &	37.3 & 38.41 &	37.54
\\
DISPOSITION& 27.53 & 31.8 &	36.7 & 34.54 & 19.75 & 25.78 &	35.87 & 20.95 & 27.54
\\
PASTSURGICAL&	29.79 &	30.07 &	36.7 & 36 & 25.76 &	31.65 &	35.87 & 32.87 & 34.12
\\
PASTMEDICALHX&	33.35 &	33.74 &	32.99 &	34.35 &	28.49 &	35.59 &	33.64 &	30.61 &	33.68
\\
ROS&	35.69 &	37.34 &	25.95 &	34.39 &	33.57 &	36.51 &	35.34 &	34.57 &	34.92
\\
GENHX&	45.63 &	45.03 &	39.13 &	44.27 &	42.72 &	41.11 &	41.41 &	39.33 &	43.41
\\
ALLERGY	& 25.01 & 22.78 & 27.26 & 8.58 & 4.48 & 44.62 &	45.68 &	43.33 &	13.11
\\
MEDICATIONS& 38.37 & 22.72 & 37.19 & 35.89 & 28.32 & 40.26 &	39.72 &	30.95 &	41.58
\\
FAM SOCHX&	33.66 &	34.43 &	32.61 &	27.17 &	28.89 &	33.74 &	27.87 & 27.45 & 33.04
\\
DIAGNOSIS&	31.54 &	35.48 &	32.61 &	34.86 &	29.2 & 52.42 &	50.33 & 47.7	& 44.94
\\
CC	& 30.4	& 28.36 & 33.24 & 30.07 & 26.25 & 31.91 & 32.54 &	34.67 &	39.14
\\
EXAM & 30.76 &	23.21 &	25.04 &	19.63 &	13.52 &	29.61 &	31.56 &30.33 & 27.97
\\
\hline

Overall	& 33.13	& 31.84 & 32.63 & 31.13 & 28.94 & 35.27 & 35.57	 & 32.68 &	32.68
\\

\hline
\end{tabular}
}
\caption{Different sections' performance across different prompting groups for GPT4. 
This is the METEOR, UMLS-F1 full table for Table \ref{table:APO_results}, and Table \ref{table:HIL_results}
.} 

\label{table:full_results4}
\end{table*}

\onecolumn
\subsection{Prompts}\label{prompts}
\lstset{
    basicstyle=\ttfamily,
    columns=fullflexible,
    breaklines=true,
    postbreak=\mbox{\textcolor{red}{$\hookrightarrow$}\space}
}
{\tiny
\centering
\begin{tabular}{|l|p{15cm}|}
\hline
\textbf{Type}  & \textbf{Prompt} \\ \hline
``Forward Pass'' &
\begin{lstlisting}
[Initial generic prompt or prompt iterations]

SOAP note section:
[section]
Conversation snippet:
[Conversation snippet]

Output your summary.
Return the output as a dictionary object, adhering to the following structure:
{"summary": ...}
Please provide your response solely in the dictionary format without including any additional text.
\end{lstlisting}
\\ \hline
$p_0$ & 
\begin{lstlisting}
In this task, we ask for your expertise in writing SOAP notes from the doctor-patient conversation.
Mainly we provide the target section in the SOAP note and the conversation snippet.
We need you to generate a summary for the respective snippet.
\end{lstlisting}
\\ \hline
$p_\nabla$ &
\begin{lstlisting}
In this task, you need to provide suggestions to modify the instruction in our SOAP notes writing system, which uses a model to generate SOAP notes from the doctor-patient conversation according to manually created instructions.
Specifically, we feed the AI a conversation snippet and the target section in the SOAP note and ask it to generate the corresponding summary.
But we found that the instruction in the current system is not perfect, so we need you to modify the instruction for this model to improve our system.

The instruction now in our rating system:
[Intial generic prompt or prompt iterations]
SOAP note section for summary:
[section]
Conversation snippet for the model:
[Conv_snippet]
Current AI summary:
[AI_summary]
Reference summary:
[label_summary]

Here are some of the requirements you need to be aware of when suggesting the instruction modification in our system:
1) For better generalization, what you suggest should be abstracted as high-level criteria as much as possible instead of only describing the details
2) We will improve the instructions based on your suggestions. If I re-provide the system with the conversation snippet and the target section in the SOAP note, it needs to be able to generate the reference summary using your new suggested instructions.
3) The instruction now in our system is for the zero-shot setting, don't try to add any examples to the instruction.
4) We are currently only focusing on this target section, so you don't need to consider the situation of other sections in the SOAP note, just optimize the instructions completely for this section.

Let's think step by step. First, output your reasons for why the current instruction in the system cannot generate the correct reference summary, then output your suggestions to modify the instruction for our system.
Return the output as a dictionary object, adhering to the following structure:
{"reasons": ..., "suggestions": ...}
Ensure the 'suggestions' only includes text but not a list. Please provide your response solely in the dictionary format without including any additional text.
\end{lstlisting}
\\ \hline
$p\delta$  & 
\begin{lstlisting}
In this task, you need to provide suggestions to modify the instruction in our SOAP notes writing system, which uses a model to generate SOAP notes from the doctor-patient conversation according to manually created instructions.
Specifically, we feed the AI a conversation snippet and the target section in the SOAP note and ask it to generate the corresponding summary.
But we found that the instruction in the current system is not perfect, so we need you to modify the instruction for this model to improve our system.

The instruction now in our system:
[Intial generic prompt or prompt iterations]
Suggestions from summary [i]:
[suggestions]
Here are some of the requirements you need to be aware of when modifying the instruction in our system:
1) For better generalization, what you suggest should be abstracted as high-level criteria as much as possible instead of only describing the details
2) We will improve the instructions based on your suggestions. If I re-provide the system with the conversation snippet and the target section in the SOAP note, it needs to be able to generate the reference summary using your new suggested instructions.
3) The instruction now in our system is for the zero-shot setting, don't try to add any examples to the instruction.
4) We are currently only focusing on this target section, so you don't need to consider the situation of other sections in the SOAP note, just optimize the instructions completely for this section.

Let's think step by step. First, briefly summarize the suggestions of all the data to get a final suggestion containing only the highest priority requirement, then output your modified instruction for our system based on the final suggestion.
Return the output as a dictionary object, adhering to the following structure:
{"final suggestion": ..., "new instruction": ...}
Please provide your response solely in the dictionary format without including any additional text.
\end{lstlisting}
\\ \hline
\end{tabular}
}
\small
Table 10: All prompts used in our proposed algorithm.

\onecolumn
\subsection{APO Iterations Examples}\label{prompts}
\lstset{
    basicstyle=\ttfamily,
    columns=fullflexible,
    breaklines=true,
    postbreak=\mbox{\textcolor{red}{$\hookrightarrow$}\space}
}
{\tiny
\centering
\begin{tabular}{|l|p{13cm}|}
\hline
\textbf{Scores}  & \textbf{Suggestions \& Prompt} \\ \hline
Initial:\\
summary\_rouge1 0.1041\\
summary\_rouge2 0.0085\\
summary\_rougeL 0.1041\\
summary\_meteor 0.0926 &
\begin{lstlisting}
In this task, we ask for your expertise in writing SOAP notes from the doctor-patient conversation.
Mainly we provide the target section in the SOAP note and the conversation snippet.
We need you to generate a summary for the respective snippet
\end{lstlisting}
\\ \hline
First epoch (good direction):\\
summary\_rouge1 0.1674\\
summary\_rouge2 0.0462\\
summary\_rougeL 0.151\\
summary\_meteor 0.2417\\
& 
\begin{lstlisting}
final suggestion in this step:  The instruction should ask the model to generate a concise summary of the patient's chief complaint based on the doctor-patient dialogue. The summary should focus on the main reason for the visit, excluding unnecessary details unless they are directly relevant to the chief complaint. The use of medical terminology should be appropriate and not overused, with the focus being on clear and accurate communication.

after curr epoch, the target_trainable_instruction is:
Based on the doctor-patient dialogue, please generate a concise summary of the patient's chief complaint, which is the primary reason for their visit to the doctor. The summary should be straightforward and focus on the main issue, excluding any unnecessary details unless they are directly relevant to the chief complaint. Use medical terminology appropriately, but ensure the summary remains clear and accurate. Remember, you are a smart doctor and a professional medical scribe, your goal is to accurately reflect the conversation.


\end{lstlisting}
\\ \hline
Second epoch (good direction):\\
summary\_rouge1 0.2079\\
summary\_rouge2 0.0704\\
summary\_rougeL 0.1897\\
summary\_meteor 0.2376 &
\begin{lstlisting}
final suggestion in this step:  The summary should be a succinct reflection of the patient's chief complaint based on the doctor-patient dialogue. It should focus solely on the main issue, excluding any details that do not directly contribute to understanding the chief complaint. The summary should be as brief as possible without losing the essence of the complaint. Use standard medical terminology appropriately, but ensure the summary remains clear and accurate. The goal is to distill the conversation into the most brief, yet comprehensive summary possible, using appropriate medical language.

after curr epoch, the target_trainable_instruction is:
Please generate a succinct summary of the patient's chief complaint, which is the primary reason for their visit, based on the doctor-patient dialogue. The summary should focus solely on the main issue, excluding any details that do not directly contribute to understanding the chief complaint. While using medical terminology appropriately, ensure the summary is as brief as possible, without losing the essence of the complaint. The goal is to distill the conversation into the most brief, yet comprehensive summary possible, using appropriate medical language. Remember, you are a smart doctor and a professional medical scribe, your task is to accurately and succinctly reflect the conversation.

\end{lstlisting}
\\ \hline
Third epoch (good direction):\\
summary\_rouge1 0.2224\\
summary\_rouge2 0.0927\\
summary\_rougeL 0.2051\\
summary\_meteor 0.255  & 
\begin{lstlisting}
final suggestion in this step:  The summary should be a concise reflection of the patient's chief complaint, capturing the essence of the doctor's inquiries and the patient's responses. It should focus on the main issue, excluding any extraneous details that do not directly contribute to understanding the chief complaint. The use of standard medical terminology is crucial, but the summary should remain brief and comprehensive. The goal is to distill the conversation into a succinct summary that accurately reflects the doctor's understanding of the chief complaint.

after curr epoch, the target_trainable_instruction is:
Please generate a concise summary of the patient's chief complaint, which is the primary reason for their visit, based on the doctor-patient dialogue. The summary should capture the essence of the doctor's inquiries and the patient's responses, focusing solely on the main issue and excluding any details that do not directly contribute to understanding the chief complaint. Use standard medical terminology to describe the complaint, ensuring the summary is as brief as possible, without losing the essence of the complaint. The goal is to distill the conversation into the most brief, yet comprehensive summary possible, using appropriate medical language. Remember, you are a smart doctor and a professional medical scribe, your task is to accurately and succinctly reflect the conversation using standard medical terms.

\end{lstlisting}
\\ \hline

Third epoch (bad direction):\\
summary\_rouge1 0.1851\\
summary\_rouge2 0.0595\\
summary\_rougeL 0.1715\\
summary\_meteor 0.2537  & 
\begin{lstlisting}
final suggestion in this step:  The instruction should guide the model to generate a concise summary of the patient's chief complaint based on the doctor-patient dialogue, focusing on the main issue and excluding irrelevant details. The summary should reflect the level of certainty or uncertainty expressed in the conversation, and accurately represent any symptoms or conditions the patient denies experiencing. The use of standard medical terminology is important, but it should not lead to verbosity. The summary should be written from the doctor's perspective, reflecting the doctor's role in the patient's care.

after curr epoch, the target_trainable_instruction is:
Please generate a concise summary of the patient's chief complaint, which is the primary reason for their visit, based on the doctor-patient dialogue. The summary should capture the essence of the doctor's inquiries and the patient's responses, focusing solely on the main issue. Exclude any details that do not directly contribute to understanding the chief complaint. Reflect the level of certainty or uncertainty expressed in the conversation. If the patient denies experiencing certain symptoms or conditions, ensure to reflect this accurately in the summary. Use standard medical terminology to describe the complaint, ensuring the summary is as brief as possible, without losing the essence of the complaint. Avoid verbosity in the use of medical terminology. The summary should be written from the doctor's perspective, reflecting the doctor's role in the patient's care. The goal is to distill the conversation into the most brief, yet comprehensive summary possible, using appropriate medical language. Remember, you are a smart doctor and a professional medical scribe, your task is to accurately and succinctly reflect the conversation using standard medical terms.

\end{lstlisting}
\\ \hline

\end{tabular}
}
\small
Table 11: APO iterations of good and bad examples from the `CC' section.

\twocolumn

\subsection{GPT Variants Per Section}\label{variants}
{\tiny
\begin{flushleft}
\begin{tabular}{|l|l|l|l|}
\hline
    \textbf{Section} & \textbf{Variant} & \textbf{Average} & \textbf{Best Variant} \\ \hline
    MEDICATIONS & text-ada-001 & 0.02255639098 & text-davinci-003 \\ \hline
    MEDICATIONS & text-babbage-001 & 0.1096938776 & text-davinci-003 \\ \hline
    MEDICATIONS & text-curie-001 & 0.09467405383 & text-davinci-003 \\ \hline
    MEDICATIONS & text-davinci-003 & 0.2071920384 & text-davinci-003 \\ \hline
    MEDICATIONS & gpt-3.5-turbo-0613 & 0.2035366419 & text-davinci-003 \\ \hline
    MEDICATIONS & gpt-4 & 0.1999162675 & text-davinci-003 \\ \hline
    PASTSURGICAL & text-ada-001 & 0.03455261137 & gpt-3.5-turbo-0613 \\ \hline
    PASTSURGICAL & text-babbage-001 & 0.02777777778 & gpt-3.5-turbo-0613 \\ \hline
    PASTSURGICAL & text-curie-001 & 0.08775603992 & gpt-3.5-turbo-0613 \\ \hline
    PASTSURGICAL & text-davinci-003 & 0.1024338849 & gpt-3.5-turbo-0613 \\ \hline
    PASTSURGICAL & gpt-3.5-turbo-0613 & 0.1309354758 & gpt-3.5-turbo-0613 \\ \hline
    PASTSURGICAL & gpt-4 & 0.1283720208 & gpt-3.5-turbo-0613 \\ \hline
    ALLERGY & text-ada-001 & 0.04682662539 & gpt-4 \\ \hline
    ALLERGY & text-babbage-001 & 0 & gpt-4 \\ \hline
    ALLERGY & text-curie-001 & 0.1891025641 & gpt-4 \\ \hline
    ALLERGY & text-davinci-003 & 0.1002458291 & gpt-4 \\ \hline
    ALLERGY & gpt-3.5-turbo-0613 & 0.2307379782 & gpt-4 \\ \hline
    ALLERGY & gpt-4 & 0.2795421063 & gpt-4 \\ \hline
    FAM/SOCHX & text-ada-001 & 0.02921216026 & gpt-4 \\ \hline
    FAM/SOCHX & text-babbage-001 & 0.03212721942 & gpt-4 \\ \hline
    FAM/SOCHX & text-curie-001 & 0.1216424461 & gpt-4 \\ \hline
    FAM/SOCHX & text-davinci-003 & 0.1441214133 & gpt-4 \\ \hline
    FAM/SOCHX & gpt-3.5-turbo-0613 & 0.2415016373 & gpt-4 \\ \hline
    FAM/SOCHX & gpt-4 & 0.26145789 & gpt-4 \\ \hline
    ASSESSMENT & text-ada-001 & 0.0388869863 & text-curie-001 \\ \hline
    ASSESSMENT & text-babbage-001 & 0.005281690141 & text-curie-001 \\ \hline
    ASSESSMENT & text-curie-001 & 0.1543199765 & text-curie-001 \\ \hline
    ASSESSMENT & text-davinci-003 & 0.1242746478 & text-curie-001 \\ \hline
    ASSESSMENT & gpt-3.5-turbo-0613 & 0.106788819 & text-curie-001 \\ \hline
    ASSESSMENT & gpt-4 & 0.1281340914 & text-curie-001 \\ \hline
    CC & text-ada-001 & 0.03660714286 & gpt-4 \\ \hline
    CC & text-babbage-001 & 0 & gpt-4 \\ \hline
    CC & text-curie-001 & 0.1886569845 & gpt-4 \\ \hline
    CC & text-davinci-003 & 0.2283677945 & gpt-4 \\ \hline
    CC & gpt-3.5-turbo-0613 & 0.2139382547 & gpt-4 \\ \hline
    CC & gpt-4 & 0.2475876016 & gpt-4 \\ \hline
    EXAM & text-ada-001 & 0.08333333333 & text-curie-001 \\ \hline
    EXAM & text-babbage-001 & 0 & text-curie-001 \\ \hline
    EXAM & text-curie-001 & 0.2142857143 & text-curie-001 \\ \hline
    EXAM & text-davinci-003 & 0.08333333333 & text-curie-001 \\ \hline
    EXAM & gpt-3.5-turbo-0613 & 0.15 & text-curie-001 \\ \hline
    EXAM & gpt-4 & 0.18 & text-curie-001 \\ \hline
    EDCOURSE & text-ada-001 & 0.1304407442 & text-davinci-003 \\ \hline
    EDCOURSE & text-babbage-001 & 0.02094356261 & text-davinci-003 \\ \hline
    EDCOURSE & text-curie-001 & 0.1772495791 & text-davinci-003 \\ \hline
    EDCOURSE & text-davinci-003 & 0.2750014022 & text-davinci-003 \\ \hline
    EDCOURSE & gpt-3.5-turbo-0613 & 0.2590712521 & text-davinci-003 \\ \hline
    EDCOURSE & gpt-4 & 0.2440284049 & text-davinci-003 \\ \hline
    ROS & text-ada-001 & 0.03748626835 & gpt-4 \\ \hline
    ROS & text-babbage-001 & 0.0340848458 & gpt-4 \\ \hline
    ROS & text-curie-001 & 0.08547537401 & gpt-4 \\ \hline
    ROS & text-davinci-003 & 0.0952141002 & gpt-4 \\ \hline
    ROS & gpt-3.5-turbo-0613 & 0.1714490651 & gpt-4 \\ \hline
    ROS & gpt-4 & 0.1762812153 & gpt-4 \\ \hline
    DISPOSITION & text-ada-001 & 0 & gpt-3.5-turbo-0613/gpt-4 \\ \hline
    DISPOSITION & text-babbage-001 & 0.1584821429 & gpt-3.5-turbo-0613/gpt-4 \\ \hline
    DISPOSITION & text-curie-001 & 0.2519607843 & gpt-3.5-turbo-0613/gpt-4 \\ \hline
    DISPOSITION & text-davinci-003 & 0.2091346154 & gpt-3.5-turbo-0613/gpt-4 \\ \hline
    DISPOSITION & gpt-3.5-turbo-0613 & 0.2608359133 & gpt-3.5-turbo-0613/gpt-4 \\ \hline
    DISPOSITION & gpt-4 & 0.2608359133 & gpt-3.5-turbo-0613/gpt-4 \\ \hline
    DIAGNOSIS & text-ada-001 & 0.05555555556 & gpt-3.5-turbo-0613 \\ \hline
    DIAGNOSIS & text-babbage-001 & 0 & gpt-3.5-turbo-0613 \\ \hline
    DIAGNOSIS & text-curie-001 & 0.05555555556 & gpt-3.5-turbo-0613 \\ \hline
    DIAGNOSIS & text-davinci-003 & 0.2532051282 & gpt-3.5-turbo-0613 \\ \hline
    DIAGNOSIS & gpt-3.5-turbo-0613 & 0.3211143695 & gpt-3.5-turbo-0613 \\ \hline
    DIAGNOSIS & gpt-4 & 0.245994832 & gpt-3.5-turbo-0613 \\ \hline
    PASTMEDICALHX & text-ada-001 & 0 & gpt-3.5-turbo-0613 \\ \hline
    PASTMEDICALHX & text-babbage-001 & 0 & gpt-3.5-turbo-0613 \\ \hline
    PASTMEDICALHX & text-curie-001 & 0.07830882353 & gpt-3.5-turbo-0613 \\ \hline
    PASTMEDICALHX & text-davinci-003 & 0.14375 & gpt-3.5-turbo-0613 \\ \hline
    PASTMEDICALHX & gpt-3.5-turbo-0613 & 0.2317706867 & gpt-3.5-turbo-0613 \\ \hline
    PASTMEDICALHX & gpt-4 & 0.2045185666 & gpt-3.5-turbo-0613 \\ \hline
    PLAN & text-ada-001 & 0.05696640316 & gpt-4 \\ \hline
    PLAN & text-babbage-001 & 0 & gpt-4 \\ \hline
    PLAN & text-curie-001 & 0.07544836116 & gpt-4 \\ \hline
    PLAN & text-davinci-003 & 0.1067404817 & gpt-4 \\ \hline
    PLAN & gpt-3.5-turbo-0613 & 0.2096407229 & gpt-4 \\ \hline
    PLAN & gpt-4 & 0.2272458144 & gpt-4 \\ \hline
    GENHX & text-ada-001 & 0.05855827354 & gpt-4 \\ \hline
    GENHX & text-babbage-001 & 0.0200537811 & gpt-4 \\ \hline
    GENHX & text-curie-001 & 0.09488431364 & gpt-4 \\ \hline
    GENHX & text-davinci-003 & 0.1421504194 & gpt-4 \\ \hline
    GENHX & gpt-3.5-turbo-0613 & 0.3101982791 & gpt-4 \\ \hline
    GENHX & gpt-4 & 0.3141274328 & gpt-4 \\ \hline
\end{tabular}
\end{flushleft}
}
    \hspace{-1em}Table 11a: The best GPT variant for each section when using the generic prompt. Note: The \textbf{Average} column is the mean of the Rouge1, Rouge2, RougeL, and RougeLsum scores.

\begin{table}[ht] 
  \begin{flushright}
  \begin{tabular}{|l|l|} 
    \hline
    \textbf{Variant} & \textbf{Count} \\ \hline 
    text-curie-001 &  2 \\ \hline
    text-davinci-003 & 2 \\ \hline
    gpt-3.5-turbo-0613 & 3 \\ \hline
    gpt-4 & 6 \\ \hline
    gpt-3.5-turbo-0613/gpt-4 & 1 \\ \hline
  \end{tabular}
  \end{flushright}
  \small
  \begin{center}
     \hspace{4em}Table11b: The number of sections where
  \end{center}
  \begin{center}
      \hspace{5em}each variant is the best. Note: The last row
  \end{center}
  \begin{center}
      \hspace{2em} is where two variants are tied for the
  \end{center}
  \begin{center}
      \hspace{-3.5em}``Disposition'' section.
  \end{center}
\end{table}
\end{document}